\documentclass{article}

\pdfoutput=1

\usepackage{microtype}
\usepackage{graphicx}
\usepackage{subfigure}
\usepackage{booktabs} 

\usepackage{hyperref}

\usepackage[accepted]{icml2023}

\usepackage{amsmath}
\usepackage{amssymb}
\usepackage{mathtools}
\usepackage{amsthm}

\usepackage[capitalize,noabbrev]{cleveref}

\theoremstyle{plain}

\theoremstyle{definition}

\theoremstyle{remark}

\usepackage[textsize=tiny]{todonotes}

\usepackage{amsmath}
\usepackage{graphicx}
\usepackage{multicol}
\usepackage{multirow}
\usepackage{makecell}
\usepackage{csquotes}
\usepackage{url}

\newcommand{\olm}{\textsc{OLMo-7B}}
\newcommand{\gmtwo}{\textsc{Gemma-2B}}
\newcommand{\gmsev}{\textsc{Gemma-7B}}
\newcommand{\lmate}{\textsc{LLaMA3-8B}}
\newcommand{\lmsev}{\textsc{LLaMA3-70B}}
\newcommand{\mis}{\textsc{Mistral-7B}}
\newcommand{\mam}{\textsc{Mamba-2.8B}}
\newcommand{\seq}[2]{\{#1^1,\dots,#1^{#2}\}}
\newcommand{\seqt}[2][1]{[t^{#1},\dots,t^{#2}]}
\newcommand{\vocab}{\mathcal{V}}
\newcommand{\xstr}{\mathbf{X}}
\newcommand{\xtok}{\mathbf{x}_\text{tok}}
\newcommand{\xalltok}{\mathbf{T}(\xstr)}
\newcommand{\dataset}{\mathcal{D}}
\newcommand{\trainset}{\dataset_\text{train}}

\newcommand{\ystr}{\mathbf{Y}}

\definecolor{navyblue}{HTML}{091F92}
\hypersetup{
    colorlinks=true,
    allcolors=navyblue,
}

\icmltitlerunning{Improving Self Consistency in LLMs through Probabilistic Tokenization}

\begin{document}

\twocolumn[
\icmltitle{Improving Self Consistency in LLMs through Probabilistic Tokenization}



\icmlsetsymbol{equal}{*}

\begin{icmlauthorlist}
\icmlauthor{Ashutosh Sathe}{equal,yyy,comp}
\icmlauthor{Divyanshu Aggarwal}{equal,comp}
\icmlauthor{Sunayana Sitaram}{comp}
\icmlaffiliation{yyy}{IIT Bombay}
\icmlaffiliation{comp}{Microsoft Research India}
\end{icmlauthorlist}


\icmlcorrespondingauthor{Ashutosh Sathe}{absathe@cse.iitb.ac.in}

\icmlkeywords{Machine Learning, ICML}

\vskip 0.3in
]



\printAffiliationsAndNotice{\icmlEqualContribution} 

\begin{abstract}
Prior research has demonstrated noticeable performance gains through the use of probabilistic tokenizations, an approach that involves employing multiple tokenizations of the same input string during the training phase of a language model. Despite these promising findings, modern large language models (LLMs) have yet to be trained using probabilistic tokenizations. Interestingly, while the tokenizers of these contemporary LLMs have the capability to generate multiple tokenizations, this property remains underutilized.

In this work, we propose a novel method to leverage the multiple tokenization capabilities of modern LLM tokenizers, aiming to enhance the self-consistency of LLMs in reasoning tasks. Our experiments indicate that when utilizing probabilistic tokenizations, LLMs generate logically diverse reasoning paths, moving beyond mere surface-level linguistic diversity. 
We carefully study probabilistic tokenization and offer insights to explain the self consistency improvements it brings through extensive experimentation on 5 LLM families and 4 reasoning benchmarks. 
\end{abstract}

\section{Introduction}

Being able to view a problem from multiple standpoints has been shown to be correlated to higher problem solving abilities and creativity \citep{gorenflo1998multiple,wang2007mp,runco2012divergent}.
In contrast, modern large language models (LLMs) such as \mis\ \citep{jiang2023mistral}, \olm\ \citep{groeneveld2024olmo}, \mam\ \citep{gu2023mamba} etc. often view language as a \textit{unique} sequence of tokens. Byte-Pair Encoding \citep{gage1994, sennrich2016neural} is a popular tokenization method employed by many LLMs to convert a given string into a sequence of tokens. Tokens in Byte-Pair Encoding (BPE) are often  ``merges'' of smaller tokens which are also present in the vocabulary. E.g. token ``\_token'' could be a merge of tokens ``\_to'' and ``ken''. This means that there are multiple valid tokenizations of a given string depending on which (sub)tokens the tokenizer encoding function chooses to merge. Table \ref{tab:intro} illustrates the phenomenon in action.

\begin{table*}[!ht] 
    \centering
    \small
    \caption{\textbf{Multiple tokenizations of a given sentence using \mis\ BPE tokenizer.} The original input string (top row) can be tokenized into multiple possible sequences of valid tokens from the \mis\ vocabulary. ``--'' represents whitespace and ``$\rvert$'' is used to indicate token boundary. The sequence of token IDs is also presented below the tokenization.}
    \begin{tabular}{p{0.95\textwidth}}
        \toprule
        A sentence can have multiple tokenizations with the BPE or Unigram tokenizer.\\ 
        \midrule        --A$\rvert$--sentence$\rvert$--can$\rvert$--have$\rvert$--multiple$\rvert$--token$\rvert$izations$\rvert$--with$\rvert$--the$\rvert$\textcolor{red}{--}$\rvert$\textcolor{red}{BP}$\rvert$\textcolor{red}{E}$\rvert$--or$\rvert$\textcolor{blue}{--Un}$\rvert$\textcolor{blue}{i}$\rvert$\textcolor{blue}{gram}$\rvert$\textcolor{magenta}{--token}$\rvert$\textcolor{magenta}{izer}$\rvert$. \\
        \{330, 12271, 541, 506, 5166, 6029, 13809, 395, 272, \textcolor{red}{28705}, \textcolor{red}{9399}, \textcolor{red}{28749}, 442, \textcolor{blue}{935}, \textcolor{blue}{28710}, \textcolor{blue}{1596}, \textcolor{magenta}{6029}, \textcolor{magenta}{4024}, 28723\}\\
        \midrule
        --A$\rvert$--sentence$\rvert$--can$\rvert$--have$\rvert$--multiple$\rvert$--token$\rvert$izations$\rvert$--with$\rvert$--the$\rvert$\textcolor{red}{--B}$\rvert$\textcolor{red}{PE}$\rvert$--or$\rvert$\textcolor{blue}{--U}$\rvert$\textcolor{blue}{ni}$\rvert$\textcolor{blue}{gram}$\rvert$\textcolor{magenta}{--token}$\rvert$\textcolor{magenta}{ize}$\rvert$\textcolor{magenta}{r}$\rvert$.\\
        \{330, 12271, 541, 506, 5166, 6029, 13809, 395, 272, \textcolor{red}{365}, \textcolor{red}{1767}, 442, \textcolor{blue}{500}, \textcolor{blue}{3023}, \textcolor{blue}{1596}, \textcolor{magenta}{6029}, \textcolor{magenta}{653}, \textcolor{magenta}{28712}, 28723\}\\
        \midrule
        --A$\rvert$--sentence$\rvert$--can$\rvert$--have$\rvert$--multiple$\rvert$--token$\rvert$izations$\rvert$--with$\rvert$--the$\rvert$\textcolor{red}{--B}$\rvert$\textcolor{red}{PE}$\rvert$--or$\rvert$\textcolor{blue}{--Un}$\rvert$\textcolor{blue}{igr}$\rvert$\textcolor{blue}{am}$\rvert$\textcolor{magenta}{--to}$\rvert$\textcolor{magenta}{ken}$\rvert$\textcolor{magenta}{izer}$\rvert$. \\
\{330, 12271, 541, 506, 5166, 6029, 13809, 395, 272, \textcolor{red}{365}, \textcolor{red}{1767}, 442, \textcolor{blue}{935}, \textcolor{blue}{3421}, \textcolor{blue}{314}, \textcolor{magenta}{298}, \textcolor{magenta}{2314}, \textcolor{magenta}{4024}, 28723\}\\
        \bottomrule
    \end{tabular}
    \label{tab:intro}
\end{table*}

Prior work \citep{kudo2018subword, provilkov2020bpe} on ``subword regularization'' has shown that including such multiple tokenizations when training neural machine translators helps the model learn better token embeddings and become robust to noisy inputs. While modern LLMs are often trained \textit{without} subword regularization, their tokenizers (mainly BPE with byte-fallback) maintain the ability to generate multiple tokenizations for a given input string as shown in Table \ref{tab:intro}. In this work, we aim to use these multiple tokenizations to improve self consistency in LLM reasoning.

An intuitive way to improve self consistency of an LLM in a reasoning task is to generate diverse reasoning paths \citep{wang2023selfconsistency} for a given problem. This is parallel to the multi-perspective reasoning as studied in \citet{wang2007mp}. Given these multiple reasoning paths, \citet{wang2023selfconsistency} propose to select the answer produced by majority of reasoning paths as the final answer. Crucially, \citet{wang2023selfconsistency} and many works that follow \citep{aggarwal2023lets,li2024turnig, jain2024lightweight, li2024escape} rely on diversity promoting text generation techniques such as Nucleus sampling \citep{Holtzman2020The} or temperature sampling to get diversity in the reasoning paths. 

In this work, we propose to use the multiple tokenizations as a primary way to generate diverse reasoning paths. Our hypothesis is that different sequences of tokens for the same input string should naturally lead to different and diverse generations. This is advantageous since unlike prior sampling methods \citep{Holtzman2020The, hewitt2022truncation, meister2023locally}, it does not rely on the model's next-token distribution to have sufficient diversity. 
It is important to note that we need a principled approach to generate multiple tokenizations of the given string since randomly dropping $p$\% of BPE merges can lead to degradation in performance as shown by \citet{jain2023baseline}. We extend \citet{kudo2018subword} and present an approach (Section \ref{sec:probtok}) to assign likelihood to a given tokenization and sampling a tokenization proportional to its likelihood. 

\section{Probabilistic Tokenization}
\label{sec:probtok}
Given an input string $\xstr$ and an existing vocabulary $\vocab$, we want to sample $m$ different tokenizations $\seq{\xtok}{m}$ such that each $\xtok^i$ is sampled proportional to $\Pr(\xtok^i|\xstr)$. Each tokenization $\xtok^i$ is a sequence of tokens $\seqt{k_i}$ where each $t^j \in \vocab$ and decoding $\xtok^i$ using $\vocab$ gives $\xstr$ back.

\subsection{Sampling a Tokenization}

We follow \citet{kudo2018subword} and use a unigram language model to estimate $\Pr(\xtok^i|\xstr)$. This means we can write $\Pr(\xtok^i|\xstr)$ using the unigram probabilities $p(t^j)$ as,

\begin{equation}
\begin{aligned}
    \Pr(\xtok^i|\xstr) &= \prod_{j=1}^{k_i}p(t^j), \\ &\text{ where } \xtok^i = \seqt{k_i}, \sum_{t^j \in \vocab}p(t^j) = 1
\end{aligned}
    \label{eq:unigram}
\end{equation}

Given the vocabulary $\vocab$ and dataset $\trainset$ of sentences, \citet{kudo2018subword} propose using the Expectation Maximization algorithm to estimate $p(t^j)$. This EM algorithm aims to maximize the marginal likelihood over the entire dataset considering $p(t^j)$ as latent variables. If $\xalltok$ denotes all possible tokenizations of a given sentence $\xstr$, the marginal can be written as,

\begin{equation}
\mathcal{L} = \sum_{s=1}^{|\trainset|}\log\Pr(\xstr_s) = \sum_{s=1}^{|\trainset|}\log\Pr\left(\sum_{\xtok\in\xalltok}\Pr(\xtok)\right)
\end{equation}

In practice, we opted for a simple counting based method to estimate $p(t^j)$. For every document in the $\trainset$, we first obtain a tokenization $\xtok^\text{BPE}$ using the existing BPE tokenizer and simply count the total occurrences of $t^j \in \xtok^\text{BPE}$ to estimate $p(t^j)$ as $\log p(t^j) = \log (\text{counts}(t^j) / N)$. Here, $N$ indicates the total number of BPE tokens produced by the tokenizer on the entire $\trainset$. For special tokens such as beginning/end of sequence, padding or unknown tokens, we set $\log p(t^j) = 0$. We provide additional discussion on counting vs EM based estimation in Appendix \ref{app:counting_vs_em}.

Once the unigram likelihoods are estimated for the entire vocabulary, we can efficiently get $l$-best tokenizations according to $\Pr(\xtok|\xstr)$ using the Forward-DP Backward-A* algorithm \citep{nagata1994stochastic}. Following \citet{kudo2018subword}, we sample from these $l$ tokenizations as $\Pr(\xtok^i|\xstr) \sim \Pr(\xtok^i)^\alpha / \sum_{i=1}^{l} \Pr(\xtok^i)^\alpha$ with $\alpha$ as a smoothing coefficient. We can now sample $m$ tokenizations for a given $\xstr$ from the best $l\geq m$-best tokenizations.

\citet{kudo2018subword} also suggests a way to accurately sample from \textit{all} ($l\rightarrow\infty$) possible tokenizations using Forward-Filtering and Backward-Sampling algorithm \citep{scott2002bayesian}. We study effects of both $l\rightarrow\infty$ and fixed $l$ on our sampled tokenizations and downstream tasks. We find that setting $l$ to a fixed value or a value dependent on $m$ can lead to less or superficial diversity in the reasoning paths sampled. More discussion on this is presented in Appendix \ref{app:finite_vs_infinite}.

\subsection{Evaluating Models with Multiple Tokenizations}\label{sec:probtok_eval_strats}

We can use probabilistic tokenization to improve self-consistency of LLMs on various types of tasks. In this work, we are interested in specifically interested in improving self consistency of reasoning based tasks.

In these tasks, we expect the model to reason through a problem (chain-of-thought) step-by-step \citep{wei2022chain} before producing the final answer. \citet{wang2023selfconsistency} has shown that sampling diverse reasoning paths from the model and picking the most common answer greatly improves the performance. While \citet{wang2023selfconsistency} rely on sampling methods such as Nucleus sampling \citep{Holtzman2020The}, we propose to use probabilistic tokenization as a way to boost diversity even further. Sampling based methods rely on the next-token distribution to be sufficiently diverse in order to generate diverse reasoning paths. As opposed to this, probabilistic tokenization directly changes the input to the model (i.e. the sequence of tokens) which should naturally lead to diverse generations. Given the reasoning problem as an input string $\xstr$, we sample $m$ different tokenizations $\seq{\xtok}{m}$ as described above and generate $m$ reasoning paths leading to $m$ different answers as $\seq{\ystr_\text{pred}}{m}$. The final answer is selected using majority vote similar to \citet{wang2023selfconsistency}. We also report ``Oracle'' task accuracy where we consider 
$\xstr$ solved if \textit{any} one of the $\seq{\ystr_\text{pred}}{m}$ matches the gold answer.

\section{Experiments}
\label{sec:experiments}


We conduct experiments to evaluate the efficacy of the proposed probabilistic tokenization on various reasoning tasks. Our findings suggest that probabilistic tokenization can robustly improve the reasoning capabilities of many LLMs.

\subsection{Setup}

\paragraph{Probabilistic Tokenization} We use a subset of the \textsc{FineWeb} dataset \citep{penedo2024fineweb} consisting of roughly 10B tokens to estimate $p(t^j)$. For both reasoning and log-likelihood based tasks, we use $l\rightarrow\infty$ i.e. we sample $m$ different from all possible tokenizations. The ablations on fixed $l$ are presented in Appendix \ref{app:finite_vs_infinite}.

\paragraph{Language Models} We experiment with four transformer-based language model families: \olm\ \citep{groeneveld2024olmo}, \gmtwo, \gmsev\ \citep{gemmateam2024gemma}, \lmate\ \citep{llama3modelcard} and \mis\ \citep{jiang2023mistral}. We also include \mam\ \citep{gu2023mamba} as a representative non-transformer based language model.

\paragraph{Reasoning Tasks} We consider four reasoning based tasks: MATH \citep{hendrycksmath2021}, AQuA \citep{ling2017program}, GSM8k \citep{cobbe2021training} and PIQA \citep{bisk2020piqa}. For each model, we report ``Baseline'' numbers which use the standard BPE tokenization. In ``CoT + SC'' baseline, we use the chain-of-thought prompting \citep{wei2022chain} with self-consistency \citep{wang2023selfconsistency} over 64 sampled reasoning paths with standard BPE tokenization. To sample diverse reasoning paths, we set the temperature $T=0.2$ with top-$k$ ($k=64$) sampling. With ``Probabilistic Tokenization'', we use the same chain-of-thought prompt. However, we sample $m=64$ different tokenizations and use greedy decoding to generate diverse reasoning paths.


\subsection{Main Results}

\begin{table*}[!ht]
    \centering
    \small
    \caption{\textbf{Results with probabilistic tokenization on chain-of-thought based reasoning tasks.} The ``Baseline'' reports the accuracy or exact-match value with the standard BPE tokenization on that task \textit{without} any chain-of-thought prompting. All other columns report changes relative to ``Baseline''. ``Probabilistic Tokenization'' (greedy decoding) outperforms chain-of-thought with self consistency (``CoT + SC'') which uses temperature based diversity promoting sampling.  }
    \begin{tabular}{llrrrrr}
        \toprule
        Task & Model & Baseline & CoT + SC & CoT + SC & Probabilistic  & Probabilistic  \\ 
        ~ & ~ & ~ & Majority & Oracle & Majority & Oracle \\
        \midrule
        \multirow{7}{*}{\bf MATH} & \olm & 3.90 & \textbf{+45.64\%} & +54.08\% & +15.34\% & +90.77\% \\ 
        ~ & \gmtwo & 5.97 & \textbf{+22.45\%} & +85.37\% & +19.03\% & +94.97\% \\ 
        ~ & \gmsev & 10.91 & +18.97\% & +42.45\% & \textbf{+20.87\%} & +120.99\% \\ 
        ~ & \lmate & 12.99 & \textbf{+28.87\%} & +82.92\% & +17.45\% & +81.60\% \\
        ~ & \lmsev & 18.18 & +16.66\% & +81.18\% & \textbf{+17.91\%} & +83.20\% \\
        ~ & \mis & 9.35 & \textbf{+45.99\%} & +78.86\% & +20.10\% & +67.27\% \\ 
        ~ & \mam & 3.12 & -23.72\% & +83.01\% & \textbf{+18.29\%} & +216.03\% \\ 
        \midrule
        \multirow{7}{*}{\bf AQuA} & \olm & 23.08 & \textbf{+23.35\%} & +53.27\% & +18.45\% & +38.60\% \\ 
        ~ & \gmtwo & 15.38 & +15.99\% & +96.52\% & \textbf{+18.63\%} & +68.79\% \\ 
        ~ & \gmsev & 23.08 & \textbf{+19.67\%} & +37.63\% & +16.62\% & +31.63\% \\ 
        ~ & \lmate & 11.54 & \textbf{+17.68\%} & +258.46\% & +16.84\% & +242.29\% \\
        ~ & \lmsev & 30.77 & +11.25\% & +13.85\% & \textbf{+12.22\%} & +15.15\% \\ 
        ~ & \mis & 23.54 & \textbf{+19.75\%} & +67.14\% & +17.80\% & +51.44\% \\ 
        ~ & \mam & 15.38 & -7.67\% & +95.85\% & \textbf{+10.21\%} & +91.81\% \\ 
        \midrule
        \multirow{7}{*}{\bf GSM8k} & \olm & 25.71 & +12.29\% & +99.04\% & \textbf{+13.22\%} & +96.42\% \\ 
        ~ & \gmtwo & 5.91 & +3.38\% & +64.91\% & \textbf{+20.81\%} & +155.84\% \\ 
        ~ & \gmsev & 25.37 & \textbf{+20.10\%} & +89.27\% & +12.61\% & +79.90\% \\ 
        ~ & \lmate & 37.71 & \textbf{+20.53\%} & +20.03\% & +13.66\% & +21.03\% \\
        ~ & \lmsev & 55.55 & +19.15\% & +35.54\% & \textbf{+21.11\%} & +32.33\% \\
        ~ & \mis & 29.66 & +20.18\% & +87.57\% & \textbf{+20.40\%} & +86.01\% \\ 
        ~ & \mam & 3.12 & +5.45\% & +113.46\% & \textbf{+28.85\%} & +68.59\% \\ 
        \midrule
        \multirow{7}{*}{\bf PIQA} & \olm & 75.89 & +17.35\% & +31.77\% & \textbf{+20.04\%} & +31.77\% \\ 
        ~ & \gmtwo & 77.17 & +16.77\% & +22.46\% & \textbf{+20.41\%} & +22.46\% \\ 
        ~ & \gmsev & 79.89 & +18.02\% & +25.17\% & \textbf{+19.40\%} & +25.17\% \\ 
        ~ & \lmate & 80.43 & +16.26\% & +17.49\% & \textbf{+17.02\%} & +17.49\% \\
        ~ & \lmsev & 80.98 & +12.61\% & +13.67\% & \textbf{+13.67\%} & +13.67\% \\ 
        ~ & \mis & 79.35 & \textbf{+20.82\%} & +26.02\% & +16.52\% & +26.02\% \\ 
        ~ & \mam & 73.91 & +1.10\% & +27.86\% & \textbf{+1.62\%} & +27.86\% \\ 
        \midrule
        ~ & Average $\Delta$ & ~
        & +16.39\% & +64.46\% & \textbf{+17.11\%} & +71.40\% \\ 
        \bottomrule
    \end{tabular}
    \label{tab:reasoning}
\end{table*}

Table \ref{tab:reasoning} shows results when applying consistency improving methods to reasoning tasks. The best (relative) performance gains for a particular task is \textbf{bolded}. We find that average performance gain obtained from ``Probabilistic Tokenization'' is higher than the average performance gain obtained by ``CoT + SC''. Interestingly, the ``Probabilistic Oracle'' is significantly better than ``CoT + SC Oracle''. By manually inspecting the reasoning traces, we find that ``Probabilistic'' tokenization is able to produce significantly diverse reasoning paths while ``CoT + SC'' reasoning paths are often just syntactically different. This can be advantageous as it gives the model more chances to improve the ``Oracle'' accuracy. 

In our experiments, the only non-transformer based model i.e. \mam\ showed mixed results when ``CoT + SC'' was applied. While the model was able to sample diverse reasoning paths, majority of them were often wrong or incomplete. Probabilistic tokenization on the other hand was able to 
consistently outperform ``CoT + SC''.

\subsection{Error Analysis}\label{sec:error_analysis}
\renewcommand{\cellalign}{tc}
\renewcommand{\theadalign}{tc}
\begin{table*}[!ht]
    \centering
    \caption{\textbf{Comparison of reasoning paths.} ``CoT + SC'' paths often have superficial linguistic diversity while ``Probabilistic Tokenization'' paths have meaningful diversity in the reasoning paths.}
    \begin{tabular}{p{0.075\textwidth}p{0.85\textwidth}}
        \toprule
        \scriptsize{Question} & \scriptsize{Sasha notices that prices for lumber have gone up 50\% in the last few months after she bought some lumber. Since she has leftovers, she decides to sell them. She has leftover ten 2 x 4 x 10 boards that cost her \$10 each. She also has five 4 x 4 x 10 boards she bought for \$16 each. If she sells them all, how much profit does she make?} \\
        \scriptsize{Answer} & \scriptsize{90}\\
        \midrule
        \makecell{\scriptsize{CoT + SC}\\\scriptsize{Path 1}} & \scriptsize{Sasha has 10 2 x 4 x 10 boards that cost her \$10 each. So she has 10 x 10 = \$100 worth of 2 x 4 x 10 boards. She also has 5 4 x 4 x 10 boards she bought for \$16 each. So she has 5 x 16 = \$80 worth of 4 x 4 x 10 boards. In total, she has \$100 + \$80 = \$180 worth of lumber. Since prices went up 50\%, she can sell them for 1.5 times as much. 1.5 x 180 = 270. The answer is \$270.}\\
        \makecell{\scriptsize{CoT + SC}\\\scriptsize{Path 2}}& \scriptsize{Sasha has ten 2 x 4 x 10 boards that cost her \$10 each. So she has 10 x \$10 = \$100 worth of 2 x 4 x 10 boards. She also has five 4 x 4 x 10 boards she bought for \$16 each. So she has 5 x \$16 = \$80 worth of 4 x 4 x 10 boards. In total, she has \$100 + \$80 = \$180 worth of lumber. Since prices have gone up 50\%, she can sell them for 1.5 times the price she bought them for. So she can sell them for 1.5 x \$180 = \$270. The answer is \$270.}\\
        \midrule
        \makecell{\scriptsize{Probabilistic}\\\scriptsize{Path 1}}& \scriptsize{Sasha has 10 boards that cost her 10 dollars each. She also has 5 boards that cost her 16 dollars each. So she spent 10 x 10 + 5 x 16 = 100 + 80 = 180. Since prices went up 50\%, she can sell the boards for 1.5 times the price she bought them for. So she can sell them for 1.5 x 180 = 270. She makes 270 - 180 = 90 dollars profit. The answer is 90.} \\
        \makecell{\scriptsize{Probabilistic}\\\scriptsize{Path 2}} & \scriptsize{Sasha bought ten 2 x 4 x 10 boards for \$10 each. So she spent 10 x 10 = \$100. She also bought five 4 x 4 x 10 boards for \$16 each. So she spent 5 x 16 = \$80. In total, she spent \$100 + \$80 = \$180. Since prices went up 50\%, she can sell the boards for 1.5 times the price she bought them for. So she can sell the 2 x 4 x 10 boards for 1.5 x 10 = \$15 each. She can sell the 4 x 4 x 10 boards for 1.5 x 16 = \$24 each. So she can sell the 2 x 4 x 10 boards for 10 x 15 = \$150. She can sell the 4 x 4 x 10 boards for 5 x 24 = \$120. In total, she can sell the boards for \$150 + \$120 = \$270. So she makes \$270 - \$180 = \$90 profit. The answer is \$90.} \\
        \bottomrule
    \end{tabular}
    \label{tab:anecdotes}
\end{table*}

To better understand the behavior of probabilistic tokenization, we study the cases where ``CoT + SC'' and ``Probabilistic Tokenization'' predictions do not match each other.

In Table \ref{tab:anecdotes}, we show a representative example from GSM8k. The model used for generating these reasoning paths is \lmate. We find that the 2 reasoning paths sampled by ``CoT + SC'' are logically equivalent. They use the same steps to in the calculation of total selling price. Importantly, they both stop early after that and get incorrect answer due to stopping early. This is a common failure mode for ``CoT + SC'' in reasoning tasks. As opposed to this, ``Probabilistic Tokenization'' is able to generate logically different reasoning paths. The second path sampled using probabilistic tokenization individually calculates the selling prices for different boards while second path calculates the total selling price directly from cost price. Notably, the difference in prompt tokenization also changes how the currency symbol is being used. In many sequences where the model verbalizes the currency (``dollar'' as opposed to ``\$'') in the reasoning paths, we noticed that the input sequence had tokenized each digit in the value separately i.e. ``\$16'' would be tokenized as ``\$$\rvert$1$\rvert$6'' as opposed to BPE's ``\$$\rvert$16''.

\section{Conclusion}
In this work, we propose ``Probabilistic Tokenization'' as a method to improve self consistency in LLMs. We present a principled way to sample multiple tokenizations of a given string using existing BPE tokenizers of pretrained LLMs. As probabilistic tokenization is an input transformation method rather than an output manipulation method, it can be generally applied to any task. 
We find that probabilistic tokenizations offers significant and consistent gains over baseline in 4 reasoning based tasks. Our analysis shows that the primary reason for success of probabilistic tokenization on reasoning tasks is its ability to generate logically diverse reasoning paths. We hope that probabilistic tokenization can be useful for providing deeper insights into the LLM's reasoning process and true abilities (``Oracle'' choices), paving the way for more robust and interpretable AI systems. 

\section{Limitations}
\label{sec:limitations}
In order to estimate the unigram probabilities $p(t^j)$ from Equation \ref{eq:unigram}, we need access to a sufficiently large and diverse dataset of documents. While we used a sample of 10B tokens from a large scale web corpus, this may not always be the optimal choice for all the tasks. On domain specific tasks such as medical question answering or code generation, a web corpus might not be appropriate. Availability to such corpus is essential since errors in estimating $p(t^j)$ can result in suboptimal tokenizations which can hurt performance \citep{jain2023baseline}. On many of the general purpose tasks, this limitation can be addressed by making use of high quality, open source web corpora such as \textsc{FineWeb} \citep{penedo2024fineweb} or \textsc{Dolma} \citep{dolma}. Even in the cases where the actual corpus is not available but token level counts are available, our method can still be applied.

Furthermore, the model capability analysis (Appendix \ref{app:model_scale}) shows that improvements from probabilistic tokenization are greater for more capable models. If the base model is not very capable, it may not be robust to changes in tokenizations or generate superficially diverse reasoning paths. In such cases, probabilistic tokenization may even \textit{hurt} the performance rather than improving. We highlight that this limitation is also present with other methods using chain-of-thought prompting or self-consistency.




\bibliographystyle{plainnat}
\bibliography{refs}







\appendix

\section{Appendix}

\subsection{Related Work}

We compare probabilistic tokenization to works in multi view learning, tokenization based improvements in language modeling and methods improving self consistency.

\paragraph{Consistency Enhancements using Input Transforms} We use a simple 2-layer neural network as our classifier to select a class given likelihood and selected option from multiple tokenizations. In principle, one can follow the rich literature on ensemble diversity \citep{stickland2020diverse, yeo2021robustness,bo2018coteaching} to build a more sophisticated reranker for both log-likelihood and reasoning based tasks that is aware of the transformed input. \citep{hao2019visual} explores similar ideas to improve consistency of outputs in computer vision.

\paragraph{Tokenization Methods to Improve LLM Performance}
Tokenization plays a crucial role in the reasoning and domain understanding capabilities for LLMs \citep{dagan2024getting, singh2024tokenization}. The popular BPE \citep{sennrich2016neural, gage1994} and Unigram \citep{kudo2018subword} tokenizers are still being studied for better understanding \citep{zouhar2023formal}. Some recent works argue that BPE might not be an optimal tokenization method for all tasks or domains \citep{liu2023task, ali2024tokenizer}. Our work is orthogonal to these directions since we do not aim to modify the existing tokenizer and LLM in any way. 

\paragraph{Self Consistency and Diversity of Thoughts in Reasoning}
Several works improve LLM reasoning capabilities using self consistency and chain-of-thought prompting\citep{wei2022chain,kojima2022large,wang2023selfconsistency, yao2023tree}. Following this, many works improve self consistency by either changing the stopping criteria  \citep{aggarwal2023lets, li2024escape} or by designing a sophisticated voting fucntion to replace majority voting \citep{li2024turnig, jain2024lightweight}. Diversity of thoughts (reasoning paths) is also shown to be an important factor limiting LLM's reasoning ability \citep{li2023making,naik2024diversity}. Our work is relevant in this direction as it aims to improve the diversity in reasoning paths using tokenization.

\subsection{Effect of model scale}\label{app:model_scale}

\begin{table*}[!ht]
    \centering
    \caption{\textbf{Comparing effectiveness of probabilistic tokenization at various model parameter scales.} Average task performance is reported for each method with percent improvement over ``Baseline'' in the bracket.}
    \begin{tabular}{lrrrrr}
        \toprule
        Model & Baseline & CoT + SC & CoT + SC & Probabilistic  & Probabilistic \\ 
         ~ & ~ & Majority & Oracle & Majority & Oracle \\
        \midrule
        \gmtwo\ & 26.11 & 29.93{\tiny(+14.65\%)} & 43.68{\tiny(+67.31\%)} & 31.26{\tiny(+19.72\%)} & 48.43{\tiny(+85.51\%)}\\
        \gmsev\ & 34.81 & 41.49{\tiny(+19.19\%)} & 51.74{\tiny(+42.54\%)} & 40.86{\tiny(+17.38\%)} & 57.24{\tiny(+64.42\%)}\\
        \lmate\ & 35.67 & 43.10{\tiny(+20.83\%)} & 69.45{\tiny(+94.73\%)} & 41.46{\tiny(+16.24\%)} & 67.98{\tiny(+90.60\%)}\\
        \lmsev\ & 46.37 & 53.29{\tiny(+14.92\%)} & 63.09{\tiny(+36.06\%)} & 53.89{\tiny(+16.23\%)} & 63.10{\tiny(+36.09\%)}\\
        \bottomrule
    \end{tabular}
    \label{tab:model_scale}
\end{table*}

We study the effectiveness of probabilistic tokenization at various model parameter scales. As shown in Table \ref{tab:model_scale}, we find that probabilistic tokenization robustly improves the performance across many model parameters scale. We find that relative improvements are highest at 7-8B parameter range, roughly when the ``reasoning'' capabilities start to emerge \citep{brown2020language}. At smaller parameter counts, less gains could be explained by the model's weaker reasoning abilities \citep{brown2020language}. Similarly, larger models that are already capable of generating diverse reasoning paths (as evidenced by lower improvements in ``Oracle'' numbers) show comparable gains over ``Baseline'' with both ``CoT + SC'' and ``Probabilstic'' methods. 

\subsection{Counting vs EM for estimating $p(t^j)$}\label{app:counting_vs_em}
As opposed to using EM to estimate $p(t^j)$ in Equation \ref{eq:unigram}, we resort to a simpler counting based method. We acknowledge that this could result in somewhat distorted unigram probabilities but is significantly faster as it does not have to enumerate over $\xalltok$. We provide additional results (Table \ref{tab:em_vs_counting}) on applying EM to a smaller, 100M token subset to conclude that both methods perform comparably. 

\begin{table*}[!ht] 
    \centering
    \caption{\textbf{Comparing effect of using counting vs EM to estimate EM probabilities.} Average task performance is reported. Both methods perform comparably on a \gmtwo\ model.}
    \begin{tabular}{lrrrrr}
        \toprule
        \multicolumn{6}{c}{Reasoning Tasks}\\
        \midrule
        Model & Baseline & CoT + SC & CoT + SC & Probabilistic  & Probabilistic \\ 
         ~ & ~ & Majority & Oracle & Majority & Oracle \\
        \midrule
        \gmtwo\ (EM) & 26.11 & 29.13 & 38.68 & 29.26 & 45.45\\
        \gmtwo\ (Counting) & 26.11 & 27.93 & 37.68 & 29.81 & 45.45\\
        \midrule
        \multicolumn{6}{c}{Loglikelihood Tasks}\\
        \midrule
        Model & Baseline & Most Likely & Majority & Classifier & Oracle\\ \midrule
        \gmtwo\ (EM) & 39.00 & 38.13 & 41.49 & 49.69 & 54.01\\
        \gmtwo\ (Counting) & 39.00 & 37.63 & 41.58 & 51.01 & 55.56\\
        \bottomrule
    \end{tabular}
    \label{tab:em_vs_counting}
\end{table*}

\subsection{Fixed $l$ vs $l\rightarrow\infty$ for sampling tokenizations}\label{app:finite_vs_infinite}
We compare using fixed vs infinite $l$ for sampling a tokenization in Table \ref{tab:fixed_vs_infinite}. Our findings suggest that considering a fixed window of top-$l$ tokenizations may not offer sufficient diversity leading to redundant generations which explain lesser improvements in ``Oracle'' as well as ``Majority'' numbers.
\begin{table*}[!ht] 
    \centering
    \caption{\textbf{Comparing effect of using counting vs EM to estimate EM probabilities.} Average task performance is reported. Both methods perform comparably on a \gmtwo\ model.}
    \begin{tabular}{lrrrrr}
        \toprule
        \multicolumn{6}{c}{Reasoning Tasks}\\
        \midrule
        Model & Baseline & CoT + SC & CoT + SC & Probabilistic  & Probabilistic \\ 
         ~ & ~ & Majority & Oracle & Majority & Oracle \\
        \midrule
        \gmtwo\ ($l=m^2$) & 26.11 & 27.13 & 33.18 & 26.26 & 35.35\\
        \gmtwo\ ($l\rightarrow\infty$) & 26.11 & \textbf{29.93} & \textbf{43.68} & \textbf{31.26} & \textbf{48.43}\\
        \midrule
        \multicolumn{6}{c}{Loglikelihood Tasks}\\
        \midrule
        Model & Baseline & Most Likely & Majority & Classifier & Oracle\\ \midrule
        \gmtwo\ ($l=m^2$) & 39.00 & \textbf{38.98} & 39.19 & 45.16 & 46.15\\
        \gmtwo\ ($l\rightarrow\infty$)& 39.00 & 38.64 & \textbf{42.98} & \textbf{53.96} & \textbf{56.06}\\
        \bottomrule
    \end{tabular}
    \label{tab:fixed_vs_infinite}
\end{table*}


\subsection{Resources Used}
\label{sec:resources_used}
We used a single NVIDIA A100 GPU with a 64 core AMD CPU to run our inferences. The estimated total GPU hours is 600 hours. Our implementation is based on the \texttt{lm-evaluation-harness}: \url{https://github.com/EleutherAI/lm-evaluation-harness} and \texttt{sentencepiece}: \url{https://github.com/google/sentencepiece}. 
We use 7 models for our experiments, all of which are opensource. The list of model and the URL with checkpoints available and licenses are listed below:
\begin{itemize}
    \item[] \olm\ : \url{https://huggingface.co/allenai/OLMo-7B} {\bf License:} Apache-2.0
    \item[] \gmtwo\ : \url{https://huggingface.co/google/gemma-2b} {\bf License:} Gemma
    \item[] \gmsev\ : \url{https://huggingface.co/google/gemma-7b} {\bf License:} Gemma
    \item[] \lmate\ : \url{https://huggingface.co/meta-llama/Meta-Llama-3-8B} {\bf License:} llama3
    \item[] \lmsev\ : \url{https://huggingface.co/meta-llama/Meta-Llama-3-70B} {\bf License:} llama3
    \item[] \mis\ : \url{https://huggingface.co/mistralai/Mistral-7B-v0.1} {\bf License:} Apache-2.0
    \item[] \mam\ : \url{https://huggingface.co/state-spaces/mamba-2.8b-hf} {\bf License:} Apache-2.0
\end{itemize}

Subsequently we used \textsc{FineWeb} dataset available at \url{https://huggingface.co/datasets/HuggingFaceFW/fineweb} to collect frequencies of the tokens in the for probabilistic tokenization mentioned in Section \ref{sec:probtok}.

\end{document}